\documentclass[
twocolumn,
]{ceurart}

\usepackage{listings}
\usepackage{tipa}
\usepackage{tabularx}
\usepackage{mdframed}
\usepackage{caption}
\usepackage{xcolor}

\mdfdefinestyle{GreenBox}{
  linecolor=black,
  linewidth=0.5pt,
  backgroundcolor=gray!10,
  innermargin=5pt,
  outermargin=5pt,
  innertopmargin=5pt,
  innerbottommargin=5pt
}

\begin{document}

\copyrightyear{2025}
\copyrightclause{Copyright for this paper by its authors.
  Use permitted under Creative Commons License Attribution 4.0
  International (CC BY 4.0).}

\conference{CLiC-it 2025: Eleventh Italian Conference on Computational Linguistics, September 24 — 26, 2025, Cagliari, Italy}

\title{When Less Is More? Diagnosing ASR Predictions in Sardinian via Layer-Wise Decoding}

\author[1,2]{Domenico De Cristofaro}[
orcid=0009-0009-7248-4896,
email=ddecristofaro@unibz.it,
]
\address[1]{Free University of Bozen-Bolzano, Italy}
\address[2]{ALPS, Alpine Laboratory of Phonetic Sciences}

\author[1,2]{Alessandro Vietti}[%
orcid=000-0002-4166-540X,
email=avietti@unibz.it,
]

\author[3]{Marianne Pouplier}
\address[3]{LMU Munich, Germany}
\author[3]{Aleese Block}

\cortext[1]{Corresponding author.}

\begin{abstract}
  Recent studies have shown that intermediate layers in multilingual speech models often encode more phonetically accurate representations than the final output layer. In this work, we apply a layer-wise decoding strategy to a pretrained Wav2Vec2 model to investigate how phoneme-level predictions evolve across encoder layers, focusing on Campidanese Sardinian, a low-resource language. We show that truncating upper transformer layers leads to improved Phoneme Error Rates (PER), with the best performance achieved not at the final layer, but two layers earlier. Through fine-grained alignment analysis, we find that intermediate predictions better preserve segmental identity, avoid overgeneration, and reduce certain classes of phonological errors. We also introduce the notion of \textit{regressive errors}—cases where correct predictions at intermediate layers are overwritten by errors at the final layer. These regressions highlight the limitations of surface-level error metrics and reveal how deeper layers may generalize or abstract away from acoustic detail. Our findings support the use of early-layer probing as a diagnostic tool for ASR models, particularly in low-resource settings where standard evaluation metrics may fail to capture linguistically meaningful behavior.
\end{abstract}

\begin{keywords}
Speech Recognition, Low-Resourced Languages, Logit Lens, Interpretability
\end{keywords}

\maketitle

\section{Introduction}

Recent research in multilingual speech foundation models has revealed that intermediate representations often encode richer phonetic information than the final output layer. Using Logit Lens-style probing across encoder layers, studies such as \citet{shim2025languagesmultilingualspeechfoundation} and \citet{langedijk-etal-2024-decoderlens} have shown that earlier layers in transformer-based models such as Whisper yield lower Word Error Rate (WER) and Character Error Rate (CER).

Building on this line of work, we investigate whether removing upper transformer layers in a pretrained multilingual ASR model influences its phoneme-level decoding behavior. Our hypothesis is grounded in prior findings—particularly those of \citet{shim2025languagesmultilingualspeechfoundation}—which demonstrate that applying a Logit Lens probing strategy to intermediate encoder layers results in lower CER for low-resource languages unseen during training. However, this raises a crucial question: \textit{what kinds of errors are actually reduced when decoding from intermediate layers instead of the full model?} More specifically, are the mistakes made by the final layer already resolved in earlier layers? To answer this,  we perform a systematic layer-wise decoding analysis using the pretrained \texttt{facebook/wav2vec2-xlsr-53-espeak-cv-ft} model on Sardinian audio data. We progressively truncate the encoder by removing a varying number of top transformer layers before decoding. For each configuration, we decode phoneme sequences and compare the output to gold-standard phonemic transcriptions, measuring overall Phoneme Error Rate (PER) and analyzing error types (insertions, deletions, substitutions). Our Contributions:
\vspace{-0.3em}
\begin{itemize}
\item we present a phoneme-level layer-wise analysis of Wav2Vec2 on a low-resource Sardinian dataset.
\item we introduce the notion of \textit{regressive errors} in ASR layer-wise decoding.
\item we show that intermediate layers (e.g., Layer 22) yield more phonetically accurate hypotheses than the final layer.
\end{itemize}

\section{Related Works}
Interpretability has become a central concern in the analysis of deep learning models for NLP and speech, particularly when it comes to understanding how linguistic representations emerge across network layers. In ASR, probing techniques such as Singular Vector Canonical Correlation Analysis (SVCCA) \cite{raghu2017svcca} and layer-wise probing classifiers \cite{belinkov2017asrprobes} have been used to assess the presence of phonetic and phonological features in hidden representations. Amnesic probing \cite{elazar2021amnesic} further shows that linguistic properties can be selectively removed from representations, suggesting that such information is not uniformly distributed across layers. A particularly effective method for layer-wise interpretability is the logit lens \cite{nostalgebraist2020logitlens}. Early exiting strategies are grounded in the observation that intermediate layers of deep neural models often suffice for accurate predictions, allowing for more efficient computation and improved robustness \cite{kaya2019shallowdeepnetworksunderstandingmitigating, schuster2022confidentadaptivelanguagemodeling, belrose2023elicitinglatentpredictionstransformers}. More recently, this idea has been extended beyond efficiency: in interpretability research, intermediate predictions have become a powerful tool for analyzing representational dynamics. The logit lens approach \cite{nostalgebraist2020logitlens}, for example, projects hidden states into output space to visualize how predictions evolve across layers. Subsequent refinements \cite{belrose2023elicitinglatentpredictionstransformers, din2024jumpconclusionsshortcuttingtransformers} have made these projections more faithful by learning layer-specific transformations, revealing how information is incrementally constructed. While these methods have mostly been explored in the context of decoder-only language models, some recent work has adapted them to speech systems. \citet{langedijk-etal-2024-decoderlens} extend the logit lens to encoder-decoder architectures such as Whisper, while \citet{shim2025languagesmultilingualspeechfoundation} demonstrate that early-layer representations in multilingual speech models may better capture phonetic distinctions—particularly in under-represented languages.
In this work, we extend this line of research by investigating why intermediate-layer decoding leads to improved performance, and whether this strategy is truly effective for low-resource languages. Rather than using early exits purely for efficiency, we treat them as a probing tool to examine how phoneme representations emerge and evolve across layers in a multilingual speech model.

\section{Methodology}

We analyze the layer-wise phoneme decoding behavior of a pretrained multilingual ASR model, \texttt{facebook/wav2vec2-xlsr-53-espeak-cv-ft} \cite{xu2021simpleeffectivezeroshotcrosslingual}, which is a wav2vec based model fine tuned on phonemic transcriptions from the Common Voice dataset \cite{ardila2020common} using a CTC loss. The model has 25 transformer encoder layers stacked above a 7 layers of convolutional feature encoder. To probe the phonetic content across layers, we apply a truncation-based decoding strategy: for each utterance, we progressively remove $k$ transformer layers (where $k \in \{0, 1, \dots, 5\}$) and perform greedy decoding on the logits computed from the last remaining layer. This is possible because all transformer layers share the same hidden dimension, allowing the model’s final projection head to be applied to intermediate layer outputs without architectural modification. As a result, we can decode phoneme sequences from any encoder layer using the same decoding pipeline. We limit the truncation to a maximum of 5 layers removed, as further reduction leads to a substantial degradation in performance, with PER increasing sharply beyond this point, reaching over 70\% of PER at Layer 16. Decoded phoneme sequences are aligned to the gold phonemic transcriptions using a phoneme-level alignment algorithm based on \texttt{SequenceMatcher}. This allows us to categorize each prediction as a correct match (hit), substitution, insertion, or deletion. Note that insertions are rarely observed in embedding-level decoding with CTC models, as output units are selected frame-wise. Many deletion errors may instead reflect phoneme mergers or coarticulation phenomena. To quantify the impact of layer removal on ASR performance, we compute the PER at each truncation level. In addition, we track phoneme-level alignment patterns and analyze the disappearance or emergence of specific error types as the number of removed layers increases.

\subsection{Dataset}

The data used in this study consists of spontaneous speech recordings in Campidanese Sardinian, a variety spoken in the southern part of Sardinia. The recordings were collected during fieldwork as part of the DID project in the municipality of Sinnai. The dataset includes 48 short utterances produced by four native speakers (two female, two male), selected from longer recordings based on linguistic relevance and clarity. The mean duration of the utterances is approximately 4.06 seconds. All utterances were manually transcribed at the phonemic level by a trained phonetician who is also a native speaker of Campidanese. The resulting dataset provides a high-quality phonemic reference for evaluating model predictions in a low-resource, under-represented language context \cite{Virdis_1988,Mereu_2020,chizzoni-vietti-2024-towards}.

\begin{figure*}[t]
    \centering
    \includegraphics[width=0.60\textwidth]{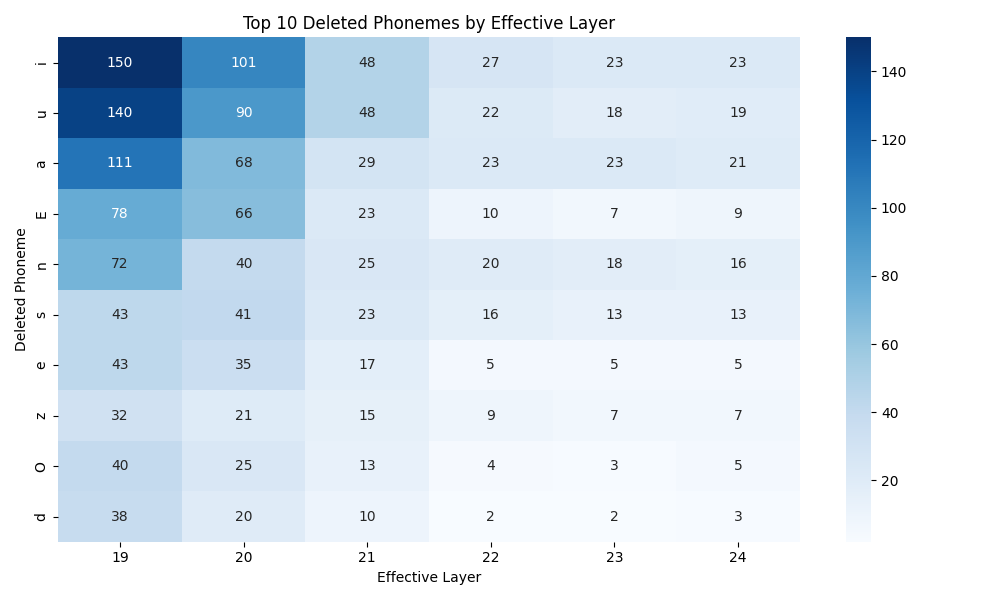}
    \hfill
    \includegraphics[width=0.60\textwidth]{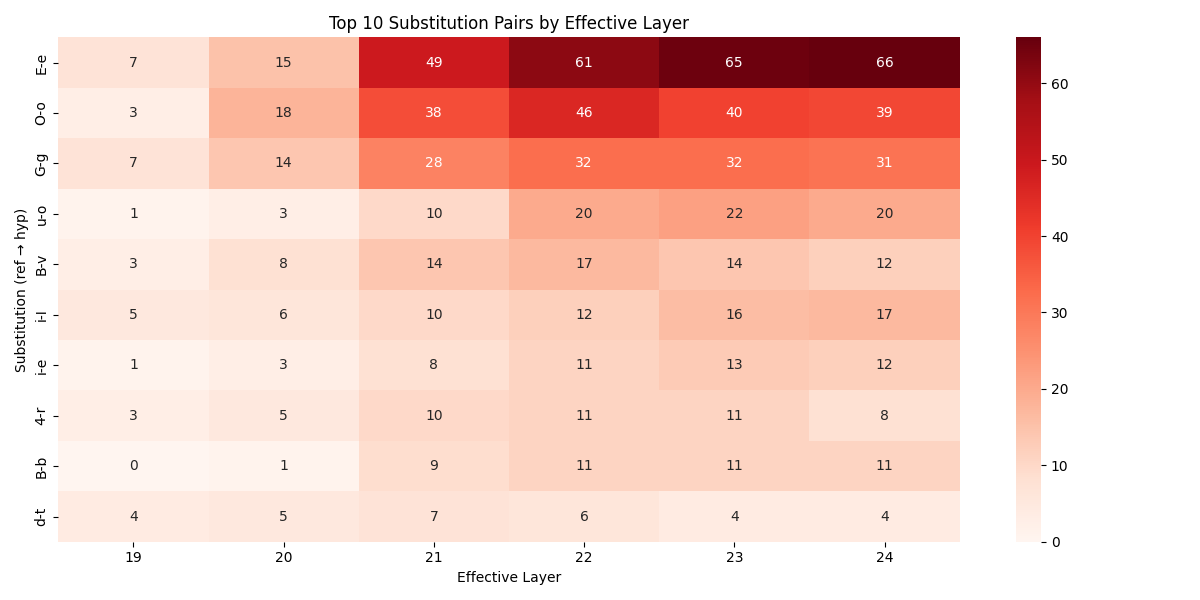}
    \caption{Heatmaps of the most frequent phoneme deletions and substitutions (ref $\rightarrow$ pred) as the number of removed transformer layers increases.}
    \label{fig:heatmaps_deletion_substitution}
\end{figure*}

\section{Results}

As shown in Table~\ref{tab:layerwise_cer}, removing the top layers of the encoder leads to a consistent reduction in PER, with the best performance observed when two layers are removed. This result supports the hypothesis that intermediate transformer layers perform better also on unseen low-resourced languages.

\begin{table}[h!]
\centering
\begin{tabular}{lc}
\toprule
\textbf{Layer} & \textbf{PER} \\
\midrule
24 & 36.73 \\
23 & 36.50 \\
22 & \textbf{35.40} \\
21 & 38.92 \\
20 & 50.03 \\
19 & 66.07 \\
\bottomrule
\end{tabular}
\caption{Phoneme Error Rate (PER) for different truncation levels.}
\label{tab:layerwise_cer}
\end{table}

\subsection{Global Trends and Error Type Evolution}

\begin{figure*}[t]
    \centering
    \includegraphics[width=0.63\textwidth]{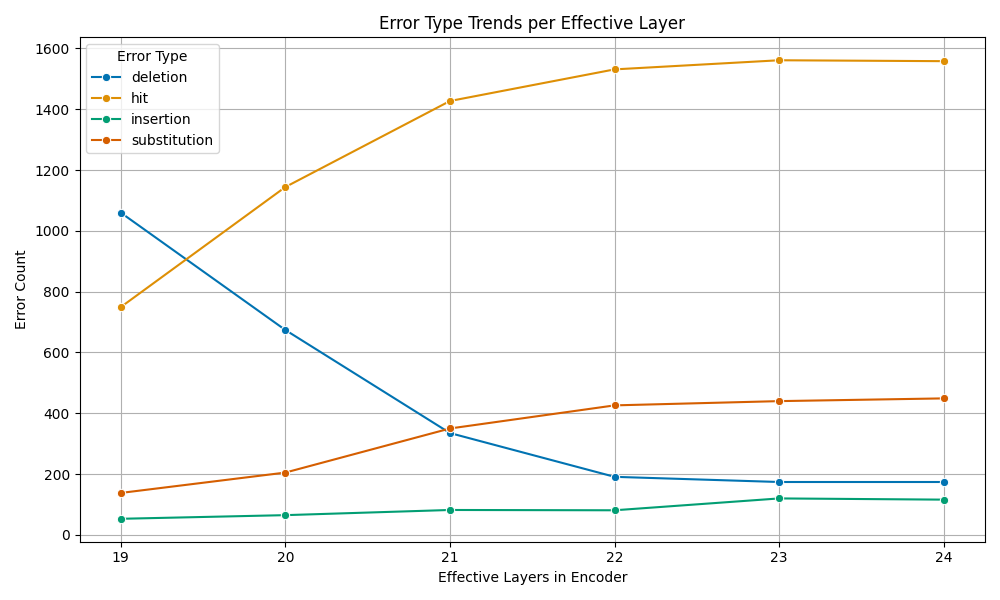}
    \caption{Trends of phoneme-level error types across effective encoder layers. While deletion errors decrease as more layers are retained, substitution errors increase. Although the number of correctly predicted phonemes (hits) also increases, it is possible that many previously deleted segments are now realized as incorrect substitutions.}
    \label{fig:error_types_trend}
\end{figure*}

Figure~\ref{fig:error_types_trend} provides a global view of how the model's phoneme-level predictions evolve as top layers are removed. As expected, the number of correctly predicted phonemes (labeled as "hit") steadily decreases as more layers are removed. At the same time, deletion errors increase sharply, particularly from Layer 21 backward, eventually dominating the error profile at Layer 19. This shows that by removing layers, the model lacks informative representations and tends to prefer skipping a prediction rather than producing an incorrect one.
In contrast, substitution errors remain relatively stable across Layers 24-22 and begin to decline slightly in deeper layers. This pattern suggests that intermediate layers may retain more accurate segment-level information, minimizing confusion between phonetically similar units. However, the sharp increase in deletions at lower layers should not be interpreted as a simple reclassification of previous substitutions. Instead, it indicates that the model is increasingly unable to resolve a segmental identity at all—maybe especially for shorter or acoustically reduced segments. At deeper layers, the model may attempt to recover some of these missing elements by assigning them a plausible phonemic category, potentially relying more on contextual or phonotactic patterns than on local acoustic evidence. This supports a view of hierarchical processing, where early layers encode fine-grained phonetic detail, while later layers abstract away from it, integrating higher-level dependencies that can both resolve and distort the original signal. However, this notion of hierarchical abstraction is model-dependent and assumes a certain architectural behavior. Since we do not impose constraints on the model design, further work is needed to test whether this abstraction emerges consistently across architectures.

To better understand these dynamics, we examine which phonemes are most frequently involved in deletion and substitution errors. As shown in Figure~\ref{fig:heatmaps_deletion_substitution}, vowel phonemes such as \textipa{/i/}, \textipa{/u/}, and \textipa{/a/} are among the most frequently deleted and substituted segments—especially as the number of removed layers increases. Interestingly, these three vowels are the only ones that commonly appear in unstressed final position in Campidanese Sardinian. While the model is not explicitly aware of word boundaries, its predictions appear sensitive to acoustic cues associated with prosodic prominence. These vowels are more likely to be reduced in duration and formant clarity when unstressed, and the model’s tendency to delete them may reflect a broader difficulty in segmenting low-prominence units—an effect we also observed in our previous analysis of stress and frequency in phoneme recognition \cite{vietti-etal-2024-sensitivity}. Some vowel deletions may also be explained by the mismatch between phoneme duration and the convolutional receptive field of the model's encoder. Since input frames are processed with overlapping windows, short vowels may be underrepresented or merged, leading to systematic omissions during decoding. Most of the substitutions involve phonetically close phoneme pairs, differing by a single articulatory feature such as voicing, manner, or vowel height. For instance, one of the most frequent substitutions is \textipa{/E/} → \textipa{/e/}, a mid-front vowel contrast distinguished primarily by height. Similarly, \textipa{/O/} → \textipa{/o/} reflects a rounded back vowel pair with a similar height difference. Another recurrent case is \textipa{/G/} → \textipa{/g/}, where a uvular fricative is replaced by a voiced plosive, suggesting the model struggles with fine-grained place and manner distinctions in lower layers. These patterns support the hypothesis that, while intermediate layers reduce substitution errors, the model's phonological representations remain coarse. Segment identity is preserved at a broad class level, but phonetic resolution weakens as contextual information is reduced. Overall, the observed substitution patterns are not random, but structured according to articulatory proximity, as further confirmed in Figure~\ref{fig:heatmaps_deletion_substitution}.

\subsection{Regressive Errors: When Hits Become Mistakes}

While final-layer predictions often improve overall accuracy, we also observe notable exceptions where the opposite occurs—cases in which the correct phoneme is already identified at an intermediate layer but becomes an error at the final layer. We refer to these as \textit{regressive errors}: instances where a phoneme is correctly predicted (a hit) at Layer 22 or 23, but turns into a substitution or deletion at Layer 24. We define a \textit{regressive error} as a case where a correct prediction (hit) at an intermediate layer~$\ell$ is replaced by a substitution or deletion at a deeper layer~$\ell + n$ (with $n > 0$). In total, we identify 53 such regressions across the dataset: 39 cases of hit $\rightarrow$ substitution and 14 cases of hit $\rightarrow$ deletion. These regressions indicate that the full encoder may in some cases “overprocess” the input, replacing a correct low-level prediction with a less accurate one as more layers are added. Crucially, most regressions involve substitutions, suggesting that deeper layers may introduce abstractions that distort fine-grained segmental information—trading off phonetic precision for contextual generalization. This may reflect a dual mechanism: (a) the re-integration of previously deleted segments, particularly those corresponding to short or hard-to-classify frames, and (b) the remapping of rare or marked phonemes onto broader, more frequent categories. In this sense, earlier layers (e.g., Layer 19) may in fact produce transcriptions that are more faithful to the phonetic input, while later layers enforce higher-level regularities at the cost of segmental detail. This challenges a common assumption: that improved overall error rates necessarily reflect more accurate linguistic representations. Instead, our findings suggest that intermediate layers may better preserve phoneme identity in certain cases, while the final layer smooths over or collapses distinctions that are phonologically relevant. To better understand the nature of these regressions, we analyze which phonemes are most frequently affected. Among the 53 cases, the high back rounded vowel \textipa{/u/} is the most common (13 instances), followed by the alveolar approximant \textipa{/r/} (7 instances), and others such as \textipa{/n/}, \textipa{/i/}, and \textipa{/a/}. Notably, many of the regressive substitutions involving \textipa{/u/} involve replacement with acoustically similar vowels like \textipa{/o/} or \textipa{/U/} in the final layer—a pattern aligned with known vowel confusions in Sardinian phonology \cite{chizzoni-vietti-2024-towards}.

\subsection{Utterances with Largest PER Reduction}
To explore whether layer truncation improves phoneme decoding in a linguistically meaningful way, we identify the five utterances that show the greatest PER reduction between Layer 24 and Layer 22 (Table~\ref{tab:sampa_predictions}). A qualitative inspection reveals that intermediate-layer outputs more closely approximate the reference transcriptions—not only in terms of segmental identity but also in overall sequence structure. While final-layer predictions sometimes exhibit phoneme insertions or reduplications that inflate the hypothesis length, the intermediate outputs tend to be more balanced and structurally coherent. This observation suggests that improvements in PER at intermediate layers are not merely an artifact of shorter sequences, but reflect more accurate segmental parsing and alignment. Rather than underpredicting, these layers appear to produce hypotheses that better capture the linguistic and prosodic shape of the input, avoiding overgeneration without compromising coverage. These improvements are quantitatively confirmed in Table~\ref{tab:per_layerwise}, where PER consistently decreases when decoding from Layer 22 compared to the full model. The most dramatic case is \texttt{03\_F\_extract\_01}, with a 50\% relative reduction in PER, followed by \texttt{30\_F\_extract\_04}, which improves by nearly 28 absolute percentage points. In both cases, the intermediate-layer output avoids spurious insertions and better aligns with the prosodic structure of the utterance. Even for more moderate improvements (e.g., \texttt{46\_M\_extract\_04} and \texttt{29\_M\_extract\_03}), we observe a shift toward more plausible segmental structures and reduced redundancy. These findings reinforce the idea that intermediate representations strike a favorable balance between acoustic faithfulness and contextual abstraction—preserving enough low-level detail to make accurate segmental decisions while avoiding the overgeneralization seen in later layers.

\begin{table*}[h!]
\centering
\small
\begin{tabular}{lrrr}
\toprule
Audio File & PER@Layer24 (\%) & PER@Layer22 (\%) & \% Improvement \\
\midrule
03\_F\_extract\_01 & 14.29 & \textbf{7.14} & 50.00 \\
30\_F\_extract\_04 & 83.33 & \textbf{55.56} & 33.33 \\
46\_M\_extract\_04 & 44.74 & \textbf{36.84} & 17.65 \\
30\_F\_extract\_02 & 46.15 & \textbf{38.46} & 16.67 \\
29\_M\_extract\_03 & 40.00 & \textbf{33.33} & 16.67 \\
\bottomrule
\end{tabular}
\caption{PER (\%) comparison between full encoder (Layer 24) and truncated model (Layer 22)}
\label{tab:per_layerwise}
\end{table*}

As illustrated in Table~\ref{tab:sampa_phoneme_error_example}, the final layer output includes several critical errors: an initial vowel \textipa{/i/} (in red) that does not appear in the reference, and an incorrect final segment /S/ (also in red) that replaces the true voiced fricative /Z/. Interestingly, at Layer 22, the model predicts a more plausible onset sequence \textipa{/e:ntsu/} (in blue), which is closer to the expected \textipa{/ensu/}, suggesting a better alignment with the reference. Additionally, the final segment \textipa{/i/} is still present in both Layer 22 and 23, but is ultimately deleted in Layer 24. This suggests that the full model may over-generalize phonetic detail, leading to the omission of segments that were correctly predicted in earlier layers. The evidence supports our broader claim: improvements in PER at intermediate layers are not merely a side-effect of over generalization, but reflect a more faithful alignment to the input acoustics. In this case, Layer 22 preserves both the segmental identity and sequence structure more reliably than the full encoder.

\begin{mdframed}[style=GreenBox, linewidth=0.8pt]
  \centering
  \begin{tabularx}{\columnwidth}{@{}lX@{}}
    \textbf{Layer 24} & \textcolor{red}{i}E5\textcolor{blue}{ntsu:}tVmla:u5Nti:\textcolor{red}{S} \\
    \textbf{Layer 23} & iE5ntsu:tamla:u5Nti:\textcolor{red}{Si:} \\
    \textbf{Layer 22} & \textcolor{blue}{e:ntsu}tamla:u5ti\textcolor{red}{Si} \\
    \textbf{\textcolor{orange}{Reference}} & ensudwamillaundiZi \\
  \end{tabularx}
  \vspace{3pt}
  \captionof{table}{Layer-wise phoneme predictions for utterance \texttt{30\_F\_extract\_04}.}
  \label{tab:sampa_phoneme_error_example}
\end{mdframed}

A similar phenomenon is observed in Table~\ref{tab:sampa_phoneme_error_example2}, where the utterance \texttt{03\_F\_extract\_01} demonstrates how the final layer introduces segmental distortions not present in earlier representations. At Layer 22, the model produces a concise and well-aligned output that accurately captures the alveolar flap /4/ (\textipa{/R/}) and avoids inserting extraneous phonetic material. Notably, the vowel preceding /4/ is realized as a short /e/ in the prediction from Layer 22, closely matching the reference transcription. In contrast, Layers 23 and 24 both produce an elongated /e:/ vowel. While this lengthening is not annotated in the reference, a manual inspection of the spectrogram reveals that the vowel is indeed phonetically long (approximately 297 ms), possibly due to prosodic or pragmatic factors. This suggests that vowel duration is a feature that only emerges at higher layers, where the model integrates broader contextual information. Rather than being an error, the elongation may reflect the model’s sensitivity to prosodic prominence, which is not explicitly captured in the phonemic gold standard but is present in the acoustic signal. In this case, then, the intermediate layer offers a segmentally accurate representation aligned with the reference, while the deeper layers introduce prosodically informed variation. This highlights how different layers may prioritize different levels of linguistic abstraction, with earlier layers preserving phonemic detail and later ones encoding broader discourse or prosodic cues.

\begin{mdframed}[style=GreenBox, linewidth=0.8pt]
  \centering
  \begin{tabularx}{\columnwidth}{@{}lX@{}}
    \textbf{Layer 24} & \textcolor{blue}{e}kambjadam\textcolor{red}{e:}4a  \\
    \textbf{Layer 23} & \textcolor{blue}{e}kambjadam\textcolor{red}{e:}4a \\
    \textbf{Layer 22} & \textcolor{blue}{e}kambjadam\textcolor{red}{e}4a\\
    \textbf{\textcolor{orange}{Reference}} & eekambjadame4a  \\
  \end{tabularx}
  \vspace{3pt}
  \captionof{table}{Layer-wise phoneme predictions for utterance \texttt{03\_F\_extract\_01}.}
  \label{tab:sampa_phoneme_error_example2}
\end{mdframed}

\section{Discussion}

Our findings challenge a widespread assumption in speech modeling: improvements in error metrics like PER necessarily reflect more accurate or linguistically meaningful predictions. While intermediate layers of the Wav2Vec2 model often yield lower PER, a closer analysis reveals that this improvement is not uniformly distributed across all phoneme classes or error types.
This aligns with an ongoing open question in speech modeling, why do higher layers often decrease WER while increasing PER? The answer may lie in how deeper layers prioritize lexical or orthographic consistency over phonetic detail, leading to better word-level predictions at the cost of segmental precision.
We observe that intermediate layers (particularly Layer 22) reduce overgeneration and avoid certain errors—such as spurious insertions or phoneme duplications—that become more frequent at deeper layers. In several cases, these intermediate predictions better align with the gold transcription both in structure and content, despite being produced with less contextual depth. Interestingly, we also identify cases of \textit{regressive errors}, where correct predictions made at intermediate layers are degraded at the final layer. These typically involve deletions or substitutions of phonemes like \textipa{/u/} and /E/, often replaced with acoustically similar segments. This suggests that deeper layers may generalize segmental contrasts. Taken together, these results indicate that error metrics like PER or CER, while useful at a high level, may obscure critical model behaviors. Intermediate representations may contain more faithful segmental information than the final output layer, particularly in under-represented or low-resource language settings. The fact that intermediate layers retain phoneme-level precision while later layers smooth over distinctions aligns with a view of hierarchical abstraction in neural models. From a phonological perspective, this might suggest that neural encoders learn generalizable phonemic categories early on and gradually shift toward context-dependent or prosodically conditioned outputs. Future work could explore whether this abstraction follows typologically consistent patterns across languages.

\section{Conclusions}

This study explored the use of layer truncation as a probing strategy for understanding phoneme-level decoding behavior in a multilingual ASR model. By applying Logit Lens-style analysis to Wav2Vec2, we show that intermediate layers can outperform the final layer in terms of Phoneme Error Rate—particularly for a low-resource language like Sardinian. Beyond aggregate improvements, our fine-grained error analysis reveals two key insights: (1) intermediate predictions tend to avoid certain types of phonological errors, and (2) in some cases, deeper layers actually degrade performance by transforming previously correct phonemes into errors. These findings suggest that the final output of a model may not always be the most linguistically faithful, especially in scenarios involving limited training data or typologically divergent phonemes. We argue that future work on speech recognition in low-resource settings should move beyond traditional evaluation metrics and incorporate layer-wise analysis as a standard interpretability tool. Doing so can provide deeper insight into how models represent phonological information—and where they fail.

\paragraph{Future work.}
While our analysis focused on Campidanese Sardinian, applying this strategy across typologically diverse low-resource languages would help determine whether the benefits of intermediate-layer decoding generalize. Additionally, attention dynamics across layers may provide further insight into which representations are retained, distorted, or lost as contextual depth increases. While the model is optimized for phoneme transcription, it is not trained on forced-aligned phoneme segmentation. Future work could investigate whether fine-tuning on time-aligned phoneme labels or segmentation tasks improves final-layer predictions and reduces regressive errors.
It would also be valuable to replicate this analysis on a language that was part of the model's pretraining or fine-tuning data (e.g., English) to assess whether intermediate layer advantages persist even in high-resource settings.

\section*{Acknowledgements}
Funded by the European Social Fund Plus Project code ESF2\_f3\_0003 “Excellence Scholarships for PhD students on topics of strategic relevance for South Tyrol”.
Work funded by the New Perspectives on Diphthong Dynamics (DID) project I83C22000390005.
\bibliography{sample-ceur}

@misc{shim2025languagesmultilingualspeechfoundation,
      title={Languages in Multilingual Speech Foundation Models Align Both Phonetically and Semantically}, 
      author={Ryan Soh-Eun Shim and Domenico De Cristofaro and Chengzhi Martin Hu and Alessandro Vietti and Barbara Plank},
      year={2025},
      eprint={2505.19606},
      archivePrefix={arXiv},
      primaryClass={cs.CL},
      url={https://arxiv.org/abs/2505.19606}, 
}

@inproceedings{langedijk-etal-2024-decoderlens,
    title = "{D}ecoder{L}ens: Layerwise Interpretation of Encoder-Decoder Transformers",
    author = "Langedijk, Anna  and
      Mohebbi, Hosein  and
      Sarti, Gabriele  and
      Zuidema, Willem  and
      Jumelet, Jaap",
    editor = "Duh, Kevin  and
      Gomez, Helena  and
      Bethard, Steven",
    booktitle = "Findings of the Association for Computational Linguistics: NAACL 2024",
    month = jun,
    year = "2024",
    address = "Mexico City, Mexico",
    publisher = "Association for Computational Linguistics",
    url = "https://aclanthology.org/2024.findings-naacl.296/",
    doi = "10.18653/v1/2024.findings-naacl.296",
    pages = "4764--4780"
}

@article{ardila2020common,
  title={Common Voice: A massively-multilingual speech corpus},
  author={Ardila, Rosana and Branson, Megan and Davis, Kelly and Kohler, Michael and Meyer, Josh and Henretty, Reuben and Morais, Michael and Saunders, Lindsay and Tyers, Francis and Weber, Gregor},
  journal={arXiv preprint arXiv:1912.06670},
  year={2020}
}

@misc{xu2021simpleeffectivezeroshotcrosslingual,
      title={Simple and Effective Zero-shot Cross-lingual Phoneme Recognition}, 
      author={Qiantong Xu and Alexei Baevski and Michael Auli},
      year={2021},
      eprint={2109.11680},
      archivePrefix={arXiv},
      primaryClass={cs.CL},
      url={https://arxiv.org/abs/2109.11680}, 
}

@inproceedings{chizzoni-vietti-2024-towards,
    title = "Towards an {ASR} System for Documenting Endangered Languages: A Preliminary Study on {S}ardinian",
    author = "Chizzoni, Ilaria  and
      Vietti, Alessandro",
    editor = "Dell'Orletta, Felice  and
      Lenci, Alessandro  and
      Montemagni, Simonetta  and
      Sprugnoli, Rachele",
    booktitle = "Proceedings of the 10th Italian Conference on Computational Linguistics (CLiC-it 2024)",
    month = dec,
    year = "2024",
    address = "Pisa, Italy",
    publisher = "CEUR Workshop Proceedings",
    url = "https://aclanthology.org/2024.clicit-1.26/",
    pages = "214--220",
    ISBN = "979-12-210-7060-6"
}

@misc{kaya2019shallowdeepnetworksunderstandingmitigating,
  title={Shallow-Deep Networks: Understanding and Mitigating Network Overthinking}, 
  author={Yigitcan Kaya and Sanghyun Hong and Tudor Dumitras},
  year={2019},
  eprint={1810.07052},
  archivePrefix={arXiv},
  primaryClass={cs.LG},
  url={https://arxiv.org/abs/1810.07052}
}

@misc{schuster2022confidentadaptivelanguagemodeling,
  title={Confident Adaptive Language Modeling}, 
  author={Tal Schuster and Adam Fisch and Jai Gupta and Mostafa Dehghani and Dara Bahri and Vinh Q. Tran and Yi Tay and Donald Metzler},
  year={2022},
  eprint={2207.07061},
  archivePrefix={arXiv},
  primaryClass={cs.CL},
  url={https://arxiv.org/abs/2207.07061}
}

@misc{nostalgebraist2020logitlens,
  author       = {nostalgebraist},
  title        = {Interpreting GPT: The Logit Lens},
  year         = {2020},
  howpublished = {\url{https://www.lesswrong.com/posts/AcKRB8wDpdaN6v6ru/interpreting-gpt-the-logit-lens}},
  note         = {Accessed: 2025-05-19}
}

@misc{belrose2023elicitinglatentpredictionstransformers,
  title={Eliciting Latent Predictions from Transformers with the Tuned Lens}, 
  author={Nora Belrose and Zach Furman and Logan Smith and Danny Halawi and Igor Ostrovsky and Lev McKinney and Stella Biderman and Jacob Steinhardt},
  year={2023},
  eprint={2303.08112},
  archivePrefix={arXiv},
  primaryClass={cs.LG},
  url={https://arxiv.org/abs/2303.08112}
}

@misc{din2024jumpconclusionsshortcuttingtransformers,
  title={Jump to Conclusions: Short-Cutting Transformers With Linear Transformations}, 
  author={Alexander Yom Din and Taelin Karidi and Leshem Choshen and Mor Geva},
  year={2024},
  eprint={2303.09435},
  archivePrefix={arXiv},
  primaryClass={cs.CL},
  url={https://arxiv.org/abs/2303.09435}
}

@inproceedings{raghu2017svcca,
  title={SVCCA: Singular vector canonical correlation analysis for deep learning dynamics and interpretability},
  author={Raghu, Maithra and Gilmer, Justin and Yosinski, Jason and Sohl-Dickstein, Jascha},
  booktitle={Advances in Neural Information Processing Systems (NeurIPS)},
  pages={6076--6085},
  year={2017}
}

@inproceedings{belinkov2017asrprobes,
  title={Analyzing phonetic and phonological knowledge in end-to-end speech recognition models},
  author={Belinkov, Yonatan and Glass, James},
  booktitle={Proceedings of the 2017 Conference on Empirical Methods in Natural Language Processing},
  pages={2396--2406},
  year={2017}
}

@misc{elazar2021amnesic,
      title={Amnesic Probing: Behavioral Explanation with Amnesic Counterfactuals}, 
      author={Yanai Elazar and Shauli Ravfogel and Alon Jacovi and Yoav Goldberg},
      year={2021},
      eprint={2006.00995},
      archivePrefix={arXiv},
      primaryClass={cs.CL},
      url={https://arxiv.org/abs/2006.00995}, 
}

@inbook{Virdis_1988, address={Tübingen}, title={Sardisch: Areallinguistik}, volume={IV-Italienisch, Korsisch, Sardisch}, booktitle={Lexikon der Romanistischen Linguistik}, publisher={Max Niemeyer}, author={Virdis, Maurizio}, editor={Holtus, Günter and Metzelin, Michael and Schmitt, Christian}, year={1988}, pages={897–913} }

@article{Mereu_2020, title={Cagliari Sardinian}, volume={50}, ISSN={0025-1003, 1475-3502}, DOI={10.1017/S0025100318000385}, abstractNote={Sardinian is a Romance language spoken almost exclusively on the island of Sardinia, an autonomous region of Italy. Sardinian and Italian are not mutually intelligible; there is considerable structural distance between the two linguistic systems, at all linguistic levels (Loporcaro 2009: 162–171).}, number={3}, journal={Journal of the International Phonetic Association}, author={Mereu, Daniela}, year={2020}, month=dec, pages={389–405}, language={en} }

@inproceedings{vietti-etal-2024-sensitivity,
    title = "Sensitivity of Syllable-Based {ASR} Predictions to Token Frequency and Lexical Stress",
    author = "Vietti, Alessandro  and
      De Cristofaro, Domenico  and
      Sara, Picciau",
    editor = "Dell'Orletta, Felice  and
      Lenci, Alessandro  and
      Montemagni, Simonetta  and
      Sprugnoli, Rachele",
    booktitle = "Proceedings of the 10th Italian Conference on Computational Linguistics (CLiC-it 2024)",
    month = dec,
    year = "2024",
    address = "Pisa, Italy",
    publisher = "CEUR Workshop Proceedings",
    url = "https://aclanthology.org/2024.clicit-1.106/",
    pages = "983--989",
    ISBN = "979-12-210-7060-6"
}

\appendix
\section{Appendix}

\begin{table}[htbp]
\centering
\scriptsize
\begin{tabularx}{\columnwidth}{lX}
\toprule
\textbf{Audio File} & \textbf{Layer-wise SAMPA Predictions} \\
\midrule
\textbf{30\_F\_extract\_04} &
24: iE5ntsu:tVmla:u5Nti:\textipa{\textesh} \newline
23: iE5ntsu:tamla:u5Nti:Si: \newline
22: e:ntsutamla:u5tiSi \newline
Ref: ensudwamillaundiZi \\
\midrule
\textbf{46\_M\_extract\_04} &
24: dedega:nivutibiStozorUnsa:kuzo:apEttsa:zU \newline
23: dedega:nivutebiStuzogunsa:kuzo:apEtsa:zu \newline
22: dedega:nivutebStzorunsakuzoapEtsazu \newline
Ref: dEdEGanivuntibiStiuzuGusakuzuapEtsauzu \\
\midrule
\textbf{30\_F\_extract\_02} &
24: snunorantazzaeti \newline
23: snunorantazzaeti \newline
22: snunorantazaeti \newline
Ref: sunO4antazEti \\
\midrule
\textbf{03\_F\_extract\_01} &
24: ekambjadame:4a \newline
23: ekambjadame:4a \newline
22: ekambjadame4a \newline
Ref: eekambjadame4a \\
\midrule
\textbf{29\_M\_extract\_03} &
L0: miza:gata:oudegonoSamuleDimiae \newline
L1: mizagata:oudegonoSamuleDimiae \newline
L2: mizagataodegonoSamuleDimiae \newline
Ref: mizEaGataudeGOnOSamullEDimiaE? \\
\bottomrule
\end{tabularx}
\caption{Layer-wise SAMPA predictions and reference for utterances with the largest PER improvement.}
\label{tab:sampa_predictions}
\end{table}

\end{document}